\documentclass[11pt]{article}

\usepackage[preprint]{acl}

\usepackage{times}
\usepackage{latexsym}
\usepackage{amssymb}
\usepackage{amsfonts}
\usepackage{amsmath}

\usepackage{dblfloatfix}

\usepackage{graphicx}
\usepackage{subcaption}

\usepackage{float}

\usepackage[T1]{fontenc}

\usepackage[utf8]{inputenc}

\usepackage{microtype}

\usepackage{inconsolata}

\usepackage{graphicx}

\usepackage{multirow}
\usepackage{booktabs}
\usepackage{csvsimple}

\title{DunbaaBERT: From Sacrifice to Semantics}

\author{
 \textbf{Iffat Maab\textsuperscript{1, $\dagger$}}
 \textbf{Waleed Jamil\textsuperscript{2}},
 \textbf{Raphael Schmitt\textsuperscript{3,4, $\dagger$}},
\\
\\
 \textsuperscript{1}Research and Development Center for Large Language Models (LLMC), National\\ Institute of Informatics, Tokyo, Japan, \textsuperscript{2}Independent Researcher, Edinburgh, United\\Kingdom,
 \textsuperscript{3}School of Computation, Information and Technology, Technical University\\ of Munich, Germany, \textsuperscript{4}Institute of General Practice, Faculty of Medicine and Medical\\ Center, University of Freiburg, Germany\\
 \textsuperscript{$\dagger$}\small{authors contributed equally}\\
 \small{
   \textbf{Correspondence:} \href{mailto:raphael.schmitt@uniklinik-freiburg.de}{raphael.schmitt@uniklinik-freiburg.de}
 }
}

\begin{document}
\maketitle
\begin{abstract}
Large language models have achieved strong performance across many NLP tasks, yet Urdu remains comparatively underexplored due to limited resources and fragmented evaluation settings. To address this gap, we introduce DunbaaBERT, a family of Urdu RoBERTa-base models trained from scratch with Byte-BPE vocabularies of 32k, 52k, and 96k tokens on a deduplicated 17\,GB Urdu corpus. We evaluate DunbaaBERT across intrinsic and downstream Urdu NLP benchmarks covering linguistic acceptability, news classification, offensive language detection, and sentiment analysis, while analyzing vocabulary-size effects on performance, pre-training dynamics, and computational efficiency. Across benchmarks, the DunbaaBERT variants achieve competitive performance against strong multilingual baselines while consistently maintaining favorable efficiency trade-offs. Under our controlled experimental setup, increasing the vocabulary size does not consistently improve downstream effectiveness, with DunbaaBERT\textsubscript{32k} repeatedly providing the strongest overall efficiency profile. Overall, our results demonstrate that carefully curated Urdu-specific encoder models can remain highly competitive despite comparatively compact model and training scales.
All models are released under the MIT license.
\end{abstract}

\section{Introduction}
Large language models have demonstrated strong performance across many NLP tasks through zero-shot, few-shot, and in-context learning approaches, often reducing the need for task-specific fine-tuning \cite{brown2020language, kojima2022large, meta2024llama, achiam2023gpt, touvron2023timothe}. However, the benefits of these methods are not distributed equally across languages. Their effectiveness is most evident in high-resource languages such as English, where large-scale pretraining data and evaluation resources are widely available. In contrast, under-resourced languages such as Urdu remain comparatively underexplored and continue to face challenges in both language understanding and generation due to limited resources \cite{maab-etal-2026-prompt} and rich morphological structure \cite{blasi2021systematic, arif2024generalists}. As a result, task-specific fine-tuning continues to play a central role in Urdu NLP, with prior work relying on supervised models for sentiment and emotion classification \cite{raza2025slur, khan-etal-2026-enhancing}, hate-speech detection \cite{arshad2023uhated, bilal2023roman, maab-etal-2026-prompt}, named entity recognition \cite{zafar-etal-2025-courtroom}, AI-generated text detection \cite{al2026u}, fake news detection \cite{dhiman2024gbert}, and more recently fact-checking \cite{ahmad-etal-2025-urdufactcheck}.

Nevertheless, compared to higher-resource languages, Urdu still lacks a broad ecosystem of openly available and systematically evaluated encoder-only language models, depsite being spoken by over 100 million people \cite{anabtawi-2026-stylometric}.
In Pakistan, Urdu remains a key language of daily communication and cultural expression, even though English continues to hold official status \cite{rahman1997urdu}. Standard Urdu is written in an abjad script historically linked to the Persian and Arabic writing traditions \cite{alam2022roman, ubul2017script}. However, online and informal communication frequently appears in Roman Urdu, where Latin characters are used to represent Urdu text \cite{10.1145/3341105.3374091}. In addition, bilingual speakers commonly alternate between Urdu and English, producing code-switched text across multiple linguistic units \citep{ansari2020use, bashir2011urdu, garcia2011urdu}. From an NLP perspective, progress in Urdu is further constrained by the scarcity and fragmentation of available resources. Existing datasets often vary in script, format, annotation scheme, and domain, spanning standard Urdu, Roman Urdu, and code-switched text \cite{saeed2021roman}. Moreover, many studies use different training and evaluation setups, which make fair comparison across methods difficult. The absence of widely adopted benchmark datasets therefore remains a major obstacle to reliable and reproducible progress in Urdu NLP \cite{maab-etal-2026-prompt}.

Recent efforts have partially improved Urdu language coverage through large-scale language-specific encoder models released within multilingual initiatives such as HPLT-BERT~\cite{samuel-etal-2023-trained, de-gibert-etal-2024-new-massive}. However, systematic analyses of Urdu-specific encoder models remain limited, particularly with respect to tokenizer design and vocabulary size. Prior work has shown that tokenization choices can substantially influence model quality in morphologically rich languages \cite{Toraman_2023}, yet such analyses remain largely unexplored for Urdu. In addition, RoBERTa-based architectures have recently demonstrated strong performance-efficiency trade-offs in Portuguese NLP \cite{scheible-schmitt-etal-2025-portbert}, motivating their investigation in the context of Urdu encoder models.

To address these limitations, we introduce DunbaaBERT, a set of Urdu RoBERTa-base models trained with Byte-BPE vocabularies of 32k, 52k, and 96k tokens. We further provide intrinsic and downstream evaluations across multiple Urdu NLP benchmarks and linguistic probing tasks, while analyzing the impact of vocabulary size on model performance, training dynamics, and efficiency trade-offs. All models are released under the MIT license to support future research in Urdu NLP.


    
    

\section{Related Works}
Multilingual pretrained encoders have played a central role in extending Transformer-based language modeling to lower-resource languages such as Urdu. mBERT~\cite{devlin-etal-2019-bert} established a strong early multilingual baseline by training a single masked language model on Wikipedia corpora from multiple languages. XLM-R~\cite{conneau-etal-2020-unsupervised} subsequently improved multilingual encoder pre-training by replacing Wikipedia-only supervision with substantially larger Common Crawl-derived multilingual data. More recently, mmBERT~\cite{marone2025mmbertmodernmultilingualencoder} has further increased multilingual scale, with the model family reported to cover over 1,800 languages and more than 3T training tokens, thereby including Urdu in a massively multilingual pre-training regime. By contrast, EuroBERT~\cite{boizard2025eurobertscalingmultilingualencoders} is explicitly designed for a restricted set of 15 European languages and is therefore not directly relevant as an Urdu baseline. While such multilingual encoders are highly important comparison points, they necessarily distribute modeling capacity across many languages and thus differ substantially from monolingual Urdu-focused pre-training.

Recent work on Urdu representation learning remains comparatively limited. One potentially relevant Hugging Face model \texttt{wajidhussain/bert-base-urdu}, is difficult to interpret reliably as an Urdu BERT baseline, since its public model card is inconsistent: it is tagged for text generation and includes GPT-2-style descriptive content rather than a standard encoder-style masked-language-model. We therefore treat it with caution and do not rely on it as a reference point.

Additional Urdu-language models hosted on Hugging Face likewise suffer from limited documentation and unclear training characteristics. The \texttt{mahwizzzz/UrduBert} model is presented as a RoBERTa-style encoder, but its model card does not provide sufficient details regarding training data, optimization setup, or pre-training duration. Preliminary experiments further suggested that the model may be substantially under-trained. Similar limitations apply to \texttt{mahwizzzz/Urdu-Bert}, whose architecture and training configuration are insufficiently documented and whose downstream behavior likewise indicates possible under-training.

Another publicly available checkpoint, \texttt{eshaaftab900/urdu-bert-base}, is described as a BERT-base model, but its model card provides only minimal technical information and lacks reproducibility-relevant details concerning corpus composition, tokenizer construction, and pre-training methodology. Due to these documentation limitations and the resulting difficulty for reliable scientific interpretation, we exclude these models from our primary comparative evaluation.

A clearer comparison point is provided by the HPLT Urdu model HPLT-BERT\textsubscript{ur}\footnote{\url{https://huggingface.co/HPLT/hplt_bert_base_ur}}, which is explicitly described as an encoder-only monolingual masked language model based on LTG-BERT, i.e., a BERT-family architecture adapted for large-scale monolingual pre-training~\cite{samuel-etal-2023-trained}. Its underlying data originate from the HPLT corpus collection~\cite{de-gibert-etal-2024-new-massive}, which was constructed from large web-derived resources including Common Crawl and Internet Archive data, with document-level deduplication applied during corpus preparation. This makes HPLT-BERT\textsubscript{ur} a relevant recent baseline for Urdu encoder pre-training on large-scale crawl-based text.

Another noteworthy line of work is RUBERT, introduced by \citet{khalid2021rubertbilingualromanurdu}, which targets Roman Urdu rather than standard Urdu written in Perso-Arabic script. RUBERT was obtained by additional pre-training of English BERT on a large Roman Urdu Twitter corpus comprising 54 million tokens and 3 million sentences. While this work demonstrates the feasibility of Urdu-adjacent pre-training in noisy social media settings, its script, domain, and initialization strategy differ substantially from monolingual encoder pre-training for standard Urdu text.
For this reason, we do not include RUBERT in our empirical comparison, as our study focuses on encoder models trained directly on Perso-Arabic Urdu corpora under controlled pre-training conditions.

\section{Methods}
\subsection{Corpora}
\label{subsec:corpora}

Our Urdu pre-training corpus was constructed from a combination of large-scale web data, encyclopedic text, and a filtered auxiliary parallel data source. The overall objective was to balance corpus size, stylistic diversity, and data quality while avoiding unnecessary complexity. To this end, we combined several complementary sources and applied targeted filtering only where the risk of noise was deemed particularly high.

\subsubsection{Core Web Corpora}

The core of our corpus consists of large-scale web-derived Urdu text. As principal sources, we used mC4 and several OSCAR subsets from CulturaX~\cite{nguyen2023culturaxcleanedenormousmultilingual}. These corpora provide broad topical and stylistic coverage and constitute the main source of linguistic variability in the final training mixture. While CulturaX represents a comparatively clean and pre-filtered corpus component, the datasets most likely still contain residual web-crawling artifacts such as boilerplate, mixed-language fragments, duplicated content, and malformed lines. Rather than aggressively filtering these sources toward a narrow stylistic target, we retained them largely in their original form to preserve domain diversity.

\subsubsection{Wikipedia}

To complement the web-derived sources with a comparatively clean and well-formed reference corpus, we included Urdu Wikipedia. The dump was converted into a one-document-per-line format, such that each line corresponds to one Wikipedia article. Compared to the web-crawled corpora, this source is more homogeneous in style and generally cleaner in terms of structure and language quality. We therefore used Wikipedia not only as an additional corpus component, but also as a reference distribution for unsupervised filtering of noisy auxiliary data. The corpus was created using WikiExtractor\footnote{https://github.com/attardi/wikiextractor}.

\subsubsection{Auxiliary Data from NLLB}

In addition to the core corpora, we considered Urdu text extracted from NLLB-derived parallel data~\cite{schwenk2020ccmatrixminingbillionshighquality, fan2020englishcentricmultilingualmachinetranslation} as an auxiliary source which we retrieved from OPUS~\cite{tiedemann-2012-parallel}. This material introduces additional diversity and potentially increases the range of lexical and syntactic constructions present in the training corpus. However, in contrast to the core corpora, this source was treated as comparatively noisy, since automatically mined multilingual or parallel resources may contain language mixing, segmentation artifacts, metadata fragments, or other irregularities. For this reason, we did not incorporate it directly, but subjected it to an additional unsupervised filtering step before inclusion.

\subsubsection{Unsupervised Filtering of Noisy Auxiliary Data}

To reduce noise in the auxiliary NLLB-derived Urdu data without relying on manual annotation, we applied an unsupervised filtering procedure based on distributional similarity to a clean reference corpus. As reference material, we used Urdu Wikipedia, which was converted into a one-document-per-line format and used to train an unsupervised fastTex skip-gram model with subword information~\cite{bojanowski2016enriching}. The resulting embedding space serves as a proxy for well-formed Urdu text.

Let $\mathcal{W} = \{w_1, \dots, w_M\}$ denote a sample of reference lines from Urdu Wikipedia. In our experiments, we set \(M = 50{,}000\), obtained from the preprocessed Wikipedia corpus in one-document-per-line format. For each \(w_i \in \mathcal{W}\), we compute a sentence representation
$\mathbf{s}_i \in \mathbb{R}^d$ using the fastText model. The reference centroid is then defined as
\[
\mathbf{r} = \frac{1}{M} \sum_{i=1}^{M} \mathbf{s}_i.
\]

Given a candidate line \(x\), we compute its sentence representation
$\mathbf{v}(x) \in \mathbb{R}^d$
and measure its similarity to the reference centroid using cosine similarity:
\[
\mathrm{sim}(x) =
\frac{\mathbf{v}(x)^{\top}\mathbf{r}}
{\|\mathbf{v}(x)\| \, \|\mathbf{r}\|}.
\]

In addition, we define a non-negative heuristic penalty term
$P(x) \geq 0,$ which captures surface-level indicators of noisy text. Concretely, \(P(x)\) is implemented as an additive penalty based on simple character-level statistics, including (i) short sequence length, (ii) low proportion of Arabic script characters, (iii) high proportion of Latin characters, digits, or punctuation, (iv) leading punctuation, (v) repeated symbols, and (vi) the presence of common boilerplate patterns. Each condition contributes a fixed penalty increment, resulting in values typically in the range \([0, 1.9]\).

The final score is given by
\[
F(x) = \mathrm{sim}(x) - P(x).
\]

A line is retained if
$F(x) \geq \tau,$
where \(\tau\) is a threshold chosen conservatively such that only clear outliers are removed. We set \(\tau=0.1\).

Applied to the NLLB-derived Urdu subset, this procedure processed a total of 18,566,353 lines, of which 17,189,655 were retained and 1,376,698 were discarded. Thus, the filtering step removed approximately 7.4\% of the candidate lines while preserving the large majority of the auxiliary material. This outcome is consistent with our goal of applying a conservative filter that removes clear outliers without collapsing the stylistic diversity of the retained text.

This approach should not be interpreted as a strict language identification system. Rather, it provides an unsupervised estimate of how closely a candidate line resembles the distribution of clean Urdu reference text. To preserve stylistic and domain diversity, we applied this filtering step only to the noisy auxiliary data rather than to the full pre-training corpus.

Appendix~\ref{sec:heuristic_penalties}, Table~\ref{tab:worst5_top5} illustrates representative high- and low-ranked candidates according to the proposed heuristic scoring function, contrasting clean sentence candidates with boilerplate and non-linguistic artifacts.



\subsubsection{Corpus Composition and Final Deduplication}
\begin{table}[t]
\centering
\small
\begin{tabular}{l r}
\hline
\textbf{Corpus} & \textbf{Size} \\
\hline
mC4 & 17.0\,GB \\
OSCAR-2019 & 869\,MB \\
OSCAR-2109 & 604\,MB \\
OSCAR-2201 & 344\,MB \\
OSCAR-2301 & 982\,MB \\
Urdu Wikipedia & 364\,MB \\
NLLB Urdu (filtered) & 2.1\,GB \\
\textit{NLLB Urdu (raw)} & \textit{2.2\,GB} \\
\hline
\textbf{Total before exact deduplication} & \textbf{22.3\,GB} \\
\textbf{Final corpus after exact deduplication} & \textbf{17.0\,GB} \\
\hline
\end{tabular}
\caption{Composition of the Urdu pre-training corpus. Sizes refer to the plain-text corpus files used during corpus construction. For NLLB, we report both the raw extracted Urdu subset and the retained subset after unsupervised filtering. The total before exact deduplication refers to the merged corpus after filtering, including the filtered NLLB portion.}
\label{tab:urdu_corpora}
\end{table}

Table~\ref{tab:urdu_corpora} summarizes the corpus components used in our experiments. The corpus primarily consists of large-scale web-derived Urdu text from mC4 and multiple OSCAR/CulturaX subsets, complemented by Urdu Wikipedia and filtered Urdu material extracted from NLLB. Together, these sources provide broad linguistic and topical diversity while balancing scale and corpus quality.

Before final deduplication, the combined corpus amounted to approximately 22.3\,GB of text. After filtering the auxiliary NLLB-derived material and merging all retained sources, we applied exact line-level deduplication, resulting in a final corpus size of approximately 17\,GB. This step removes literal duplicates across sources while preserving most of the variability of the training data. We deliberately refrained from more aggressive fuzzy or near-duplicate removal procedures, since such methods may over-regularize the resulting corpus.

In the final corpus, each line corresponds to one document. After deduplication, the corpus was randomly shuffled prior to binarization and pre-training to avoid source-local ordering effects.

\subsection{Pre-training}

DunbaaBERT was trained from scratch using our modular pre-training framework introduced in \citet{SCHMITT2026100824}, which provides a reproducible pipeline for large-scale RoBERTa-style model training and orchestrates binary dataset generation, distributed training, check-pointing, and experiment management. The framework builds on fairseq \citep{ott_fairseq_2019} as the underlying training engine.

All models followed the RoBERTa-base architecture and were trained for 100k update steps using dynamic masking with whole word masking (WWM), fixed input sequences of 512 tokens, and optimization settings closely following \citet{liu2019robertarobustlyoptimizedbert}. The use of WWM ensures that masking operates at the word level rather than on individual subword units, which is particularly relevant in the context of varying vocabulary sizes. A warmup phase of 10k updates gradually increased the learning rate to a peak of $7 \times 10^{-4}$, followed by polynomial decay to zero.

Training was conducted in mixed precision (FP16) with an effective batch size of 8k. Pre-training was performed on NVIDIA H100 GPUs (96GB), using two GPUs for DunbaaBERT\textsubscript{32k} and DunbaaBERT\textsubscript{96k}, and four GPUs for DunbaaBERT\textsubscript{52k}. Aside from the vocabulary sizes under investigation, all pre-training settings were kept identical across model variants to isolate tokenizer effects.

During pre-training, sequences were constructed while preserving natural sentence boundaries, avoiding mid-sentence truncation and maintaining linguistic structure throughout optimization.

\section{Experiments}

\begin{table*}[!b]
\centering
\resizebox{\textwidth}{!}{
\setlength{\tabcolsep}{3pt}
\begin{tabular}{lcccccccccccccccccccc}
\hline
\textbf{Model} & Asp. Agr. & DatObj N. & DatObj Pr. & Erg. VGDat. & Erg. PerfV. & Erg. ObjVAg. & Exp. Subj. & Honorific & N-Adj & Obl. SN. & Obl. Pron. & Obl. Adj. & Obl. Plur. & Obl. Verb & Part. Rel. & S-V Gender & S-V Num. & S-V Pers. & WordOrdVar & \textbf{AVG} \\
\hline
\csvreader[late after line=\\]{urdu_blimp_detail.csv}{}%
 {\csvcoli & \csvcolii & \csvcoliii & \csvcoliv & \csvcolv & \csvcolvi & \csvcolvii & \csvcolviii & \csvcolix & \csvcolx & \csvcolxi & \csvcolxii & \csvcolxiii & \csvcolxiv & \csvcolxv & \csvcolxvi & \csvcolxvii & \csvcolxviii & \csvcolxix & \csvcolxx & \csvcolxxi}
\hline
\end{tabular}
}
\caption{\label{tab:urdu_blimp_detail}
Detailed \textsc{UrBLiMP} evaluation across 19 linguistic acceptability categories. 
Best results are shown in bold and second-best results are underlined.}
\end{table*}

\subsection{Datasets}

\paragraph{Linguistic Acceptability}
To assess fine-grained grammatical knowledge, we include evaluation on \textbf{\textsc{UrBLiMP}}~\citep{adeeba2025urblimpbenchmarkevaluatinglinguistic}, a benchmark of minimal pairs designed to probe syntactic and morphosyntactic competence in Urdu. \textsc{UrBLiMP} comprises 5{,}696 minimal sentence pairs targeting ten core linguistic phenomena, including agreement, argument structure, and word order variations. Each pair contrasts a grammatical and an ungrammatical sentence differing only minimally. For our analysis, we report results on 19 fine-grained evaluation categories derived from these phenomena, enabling a more detailed assessment of model behavior across specific linguistic constructions. Models are evaluated following the BLiMP protocol~\citep{warstadt-etal-2020-blimp-benchmark}, i.e., by assigning higher probability to the grammatical sentence in each pair. We report accuracy for each category and use the macro-average across all categories as the overall \textsc{UrBLiMP} score, complementing downstream tasks by evaluating fine-grained grammatical knowledge.

\paragraph{News Domain Classification}
\textbf{COUNT19} contains 10,451 documents collected from major Urdu news websites, including Geo, ARY, Express, Samaa, and BBC Urdu \cite{9022736}. It covers seven news domains: International, National, Science, Sports, Health, Business, and Entertainment. The dataset is relatively balanced across categories, with each class containing approximately 1,480--1,509 documents, and contains 3.19M tokens, 91,840 unique unigram tokens, and 759,997 unique bigram tokens. Following the original dataset construction by \citet{9022736}, for COUNT19, we used the concatenation of the news title and article content as the input text for multi-classification among the seven categories.

\paragraph{Offensive Language Detection}
\textbf{USADC} is an Urdu-script abusive language detection dataset formulated as a binary classification task utilized by \citet{maab-etal-2026-prompt, arif-etal-2024-generalists, amjad2022overview}. The dataset contains 5,670 instances, with 2,816 labeled as \textit{Normal} and 2,854 as \textit{Abusive}. Unlike most Roman Urdu abuse detection datasets \cite{maab-etal-2026-prompt}, USADC preserves the native Urdu script, making it particularly useful for evaluating models on script-specific lexical, morphological, and orthographic patterns in Urdu abusive language detection.

\paragraph{Sentiment Classification}
Since sentiment classification is one of the most widely studied downstream tasks in Urdu NLP~\cite{ashraf2024revolutionizing, khan2022multi, sent_ashraf}, we include it as an important benchmark task. For this task, we use \textbf{PSL–Kabaddi} benchmark, an Urdu Twitter sentiment dataset derived from a collection of sports-related tweets about the Pakistan Super League (PSL) and Kabaddi \cite{maqsood-2023-weakly}. The original data were collected from trending topics in Pakistan and subsequently filtered to remove non-Urdu tweets. The benchmark contains tweet-level aspect and sentiment annotations. In our experiments, we use the sentiment polarity labels for classification. The dataset contains 2,924 sentiment labels, with 1,577 positive, 1,100 negative, and 245 neutral labels, indicating a moderately imbalanced sentiment distribution.

We also use an Urdu \textbf{IMDB} movie review dataset for document-level binary sentiment classification task\footnote{\url{https://www.kaggle.com/datasets/akkefa/imdb-dataset-of-50k-movie-translated-urdu-reviews}}. This dataset is an Urdu translation of the original English IMDB movie review corpus, produced using Google Translate, and contains 50,000 reviews evenly split between positive and negative classes; it has also been used in prior Urdu sentiment analysis studies \citep{irum2025document, arif-etal-2024-generalists, hassan2024polarity}.

\subsection{Setup}

All downstream experiments were conducted using a unified fine-tuning setup based on PyTorch and HuggingFace Transformers. We performed hyperparameter optimization (HPO) across multiple learning rates, batch sizes, and random seeds, training each configuration for up to 30 epochs with early stopping (patience = 3). Best-performing configurations were selected based on validation macro-F1 and subsequently evaluated on the test set using a fixed evaluation batch size of 8. Additional implementation details, hyperparameter ranges, hardware specifications, dataset splits, and baseline properties are provided in Appendix~\ref{sec:appendix_tech}. To characterize variability across the complete hyperparameter search space, Appendix~\ref{sec:performance_efficiency_all_60}, Table~\ref{tab:all_benchmark_normalized_efficiency}, reports results for all 60 configurations per model and benchmark, comprising five learning rates, four batch sizes, and three random seeds. Computational costs are summarized in Appendix~\ref{sec:computational_cost_of_benchmarking}.


\subsection{Evaluation and Efficiency Metrics}
\label{subsec:evaluation_metrics}

We report Accuracy, Macro-F1, and normalized efficiency (Norm. Eff.) for all downstream tasks. Within each benchmark, normalized efficiency is defined as
\[
\mathrm{Norm.\ Eff.}
=
\left(\frac{\mathrm{Macro\mbox{-}F1}}{100}\right)
\times
\left(\frac{\mathrm{SPS}}{\max(\mathrm{SPS})}\right),
\]
where SPS denotes the number of test samples processed per second using a fixed evaluation batch size of 8, and $\max(\mathrm{SPS})$ is the highest mean SPS among the evaluated models for that benchmark. Higher values indicate a stronger performance--throughput trade-off. For the all-configuration analysis, mean Macro-F1 and mean SPS across the 60 configurations were used to calculate normalized efficiency. Normalization was performed separately within each benchmark to account for differences in dataset characteristics and inference workloads. For \textsc{UrBLiMP}, we report the macro-average of category-wise accuracies.

\section{Results}

\subsection{Pre-training Dynamics}
\label{sec:pre-training_dynamics}

\begin{figure}[htb]
    \centering
    \includegraphics[width=\columnwidth]{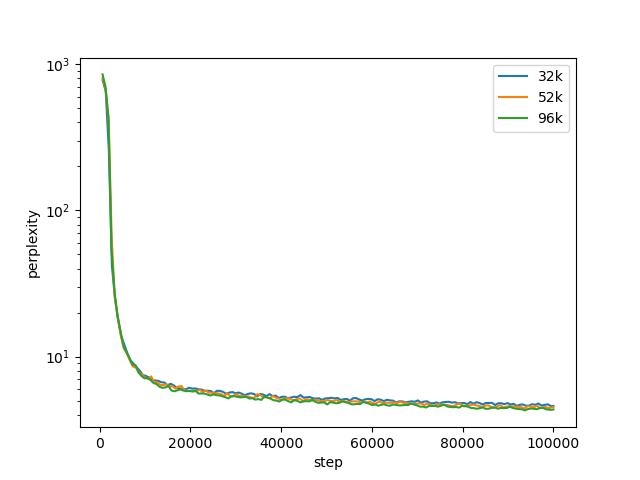}
    \caption{Validation perplexity across DunbaaBERT models with 32k, 52k, and 96k vocabularies. Nearly overlapping convergence curves and only marginal differences in final perplexity suggest limited impact of vocabulary size on intrinsic pre-training behavior.}
    \label{fig:perplexity-valid}
\end{figure}

During pre-training, perplexity was tracked on the validation set at each checkpoint (Figure \ref{fig:perplexity-valid}). Across all three DunbaaBERT variants (32k, 52k, and 96k vocabularies), perplexity decreases rapidly during the early stages of training, followed by smooth and stable convergence. After the initial steep drop, convergence becomes notably flatter, with only marginal improvements after approximately 40k--50k updates, suggesting that the models approach saturation well before the end of training.

Notably, the perplexity trajectories of all three vocabulary configurations are almost indistinguishable. While minor fluctuations can be observed, these are transient and do not indicate instability. All models exhibit highly similar convergence behavior, pointing to stable optimization and good generalization.

A particularly interesting observation is that vocabulary size appears to have only negligible influence on perplexity. Despite substantial differences between 32k, 52k, and 96k token vocabularies, the validation curves nearly overlap throughout training, with final validation perplexities differing only marginally (4.59, 4.52, and 4.35, respectively). While the 96k vocabulary achieves the numerically lowest perplexity, the absolute differences remain small, suggesting that intrinsic language modeling behavior is governed primarily by the characteristics of the pre-training corpus rather than by vocabulary size itself. In other words, increasing vocabulary size beyond 32k yields only diminishing returns in intrinsic modeling quality, motivating assessment of vocabulary choices mainly through downstream performance and computational efficiency rather than perplexity alone.

We further observe that the total number of completed epochs differs slightly across vocabulary configurations at a fixed budget of 100k update steps. The 32k, 52k, and 96k models reach approximately 147, 150, and 152 epochs, respectively. This variation is expected, as vocabulary size affects sequence packing efficiency and thus the number of effective training examples processed per update. However, despite these small differences in epoch counts, all models exhibit nearly identical convergence behavior and final perplexity values, reinforcing the observation that vocabulary size has only a minor impact on intrinsic pre-training dynamics in our setting.

\subsection{Linguistic Acceptability}
\label{sec:urblimp}

Table~\ref{tab:urdu_blimp_detail} reports detailed results on \textsc{UrBLiMP}, evaluating fine-grained syntactic and morphosyntactic competence across 19 linguistic categories. Across all DunbaaBERT variants, we observe consistently strong performance, substantially outperforming multilingual baselines such as mBERT, mmBERT, and XLM-R. The strongest overall results are achieved by HPLT-BERT\textsubscript{ur}. DunbaaBERT remains competitive with HPLT-BERT\textsubscript{ur} while using a controlled and comparatively compact pre-training setup, highlighting the effectiveness of our approach.

Focusing on the vocabulary ablation, the DunbaaBERT models exhibit a clear but nuanced pattern. DunbaaBERT\textsubscript{52k} achieves the best overall performance with an average score of 97.0, followed by DunbaaBERT\textsubscript{32k} (95.1) and DunbaaBERT\textsubscript{96k} (94.6). While all three models perform at a high level, the differences between them are more pronounced than those observed in pre-training perplexity, indicating that downstream linguistic competence is more sensitive to vocabulary design than intrinsic language modeling metrics.

Interestingly, the medium-sized vocabulary (52k) yields the strongest and most balanced results across categories, achieving top or near-top performance in multiple areas, including agreement, oblique marking, and subject–verb agreement. In contrast, the 32k model performs competitively but shows slightly weaker performance in more complex agreement and word order categories. The 96k model, despite achieving the lowest perplexity during pre-training, does not translate this advantage into improved downstream performance and instead exhibits small but consistent degradations across several categories.

Overall, under our controlled experimental setup, increasing the vocabulary size from 52k to 96k did not improve fine-grained grammatical performance. Because the DunbaaBERT variants share the same architecture, pre-training corpus, training budget, and optimization settings, these results isolate vocabulary size as the primary experimental difference. However, they should not be interpreted as establishing a universally optimal vocabulary size for Urdu.

\subsection{Downstream Tasks}

\begin{table*}[htb]
\centering
\tiny
\resizebox{\textwidth}{!}{%
\begin{tabular}{l|ccc|ccc|ccc|ccc}
\toprule
\multirow{2}{*}{\textbf{Model}}
& \multicolumn{3}{c|}{\textbf{COUNT19}}
& \multicolumn{3}{c|}{\textbf{USADC}}
& \multicolumn{3}{c|}{\textbf{PSL--Kabaddi}}
& \multicolumn{3}{c}{\textbf{IMDB Urdu}} \\
& F1 & Acc. & Norm. Eff.
& F1 & Acc. & Norm. Eff.
& F1 & Acc. & Norm. Eff.
& F1 & Acc. & Norm. Eff. \\
\midrule
DunbaaBERT\textsubscript{32k} & 94.44 & 95.17 & \textbf{0.944} & \textbf{94.08} & \textbf{94.15} & 0.893 & 70.08 & 82.51 & \textbf{0.701} & 90.13 & 90.14 & \textbf{0.897} \\
DunbaaBERT\textsubscript{52k} & 94.91 & 95.46 & 0.908 & 91.75 & 91.81 & \underline{0.895} & 67.60 & 81.40 & 0.634 & 90.14 & 90.15 & 0.744 \\
DunbaaBERT\textsubscript{96k} & \underline{95.22} & \underline{95.96} & \underline{0.917} & 89.97 & 90.06 & \textbf{0.900} & \underline{70.53} & \textbf{82.78} & \underline{0.655} & \underline{90.65} & \underline{90.66} & 0.779 \\
Urdu-RoBERTa\textsubscript{small} & 92.08 & 93.39 & 0.890 & 85.36 & 85.38 & 0.792 & 67.06 & 77.55 & 0.595 & 84.72 & 84.73 & 0.847 \\
HPLT\textsubscript{base} & \textbf{95.71} & \textbf{96.35} & 0.585 & \underline{93.51} & \underline{93.57} & 0.764 & \textbf{71.11} & 81.54 & 0.487 & 89.69 & 89.70 & 0.553 \\
mBERT & 90.88 & 92.37 & 0.825 & 83.03 & 83.04 & 0.716 & 65.78 & 76.03 & 0.587 & 85.47 & 85.47 & 0.846 \\
mmBERT\textsubscript{small} & 92.36 & 93.43 & 0.513 & 73.09 & 73.10 & 0.570 & 70.36 & 79.48 & 0.432 & 85.44 & 85.44 & 0.461 \\
mmBERT\textsubscript{base} & 93.97 & 94.71 & 0.522 & 77.77 & 77.78 & 0.582 & 67.75 & 78.37 & 0.409 & 87.31 & 87.31 & 0.468 \\
XLM-R\textsubscript{base} & 93.72 & 94.54 & 0.910 & 85.22 & 85.38 & 0.712 & 60.56 & 79.75 & 0.539 & 88.69 & 88.70 & \underline{0.855} \\
XLM-R\textsubscript{large} & 94.38 & 95.04 & 0.512 & 83.55 & 83.63 & 0.576 & 69.62 & \underline{82.64} & 0.405 & \textbf{91.15} & \textbf{91.15} & 0.476 \\
\bottomrule
\end{tabular}%
}
\caption{Results across four Urdu downstream benchmarks. Norm. Eff. denotes normalized efficiency (see \ref{subsec:efficiency_metric}). Best results are shown in bold and second-best results are underlined.}
\label{tab:all_benchmarks_summary_runtime_inference}
\end{table*}

Table~\ref{tab:all_benchmarks_summary_runtime_inference} summarizes the downstream results for news domain classification, offensive language detection, and sentiment classification. Variability across the complete hyperparameter search space is reported in Appendix~\ref{sec:performance_efficiency_all_60}, Table~\ref{tab:all_benchmark_normalized_efficiency}, which summarizes 60 configurations per model and benchmark as mean $\pm$ standard deviation. Additional search-cost and efficiency analyses are provided in Figure~\ref{fig:search_cost_efficiency_six_panel}. Overall, the DunbaaBERT variants achieve competitive task performance while consistently maintaining strong normalized efficiency across benchmarks. In contrast, larger multilingual models can achieve strong raw classification scores, but at substantially weaker efficiency trade-offs.

\paragraph{News Domain Classification}
On COUNT19, HPLT\textsubscript{base} achieves the highest Macro-F1 and accuracy scores (95.71 and 96.35), followed closely by DunbaaBERT\textsubscript{96k} (95.22 and 95.96). However, the DunbaaBERT variants achieve substantially stronger efficiency trade-offs, with DunbaaBERT\textsubscript{32k} obtaining the highest normalized efficiency score of 0.944. The broader efficiency analysis in Appendix~\ref{sec:performance_efficiency_all_60} further shows that the DunbaaBERT variants maintain favorable throughput and training-time behavior across the full hyperparameter search space.

\paragraph{Offensive Language Detection}
On USADC, DunbaaBERT\textsubscript{32k} achieves the strongest overall performance with a Macro-F1 score of 94.08 and an accuracy of 94.15. The DunbaaBERT variants also dominate normalized efficiency, with DunbaaBERT\textsubscript{96k} achieving the highest score (0.900) and DunbaaBERT\textsubscript{52k} the second-highest score (0.895). HPLT\textsubscript{base} remains competitive with a Macro-F1 score of 93.51, whereas multilingual baselines such as mmBERT and XLM-R show noticeably weaker downstream effectiveness and efficiency.

\paragraph{Sentiment Classification}
For sentiment classification, the results differ across the two benchmarks. On PSL--Kabaddi, HPLT\textsubscript{base} achieves the highest Macro-F1 score (71.11), while DunbaaBERT\textsubscript{96k} obtains the highest accuracy (82.78). Interestingly, mmBERT\textsubscript{small} also performs competitively on this benchmark, reaching a Macro-F1 score of 70.36 despite comparatively weak normalized efficiency. On IMDB Urdu, XLM-R\textsubscript{large} achieves the highest Macro-F1 and accuracy scores (91.15), indicating that larger multilingual encoders may benefit from broader semantic coverage on large-scale document-level sentiment tasks. Nevertheless, DunbaaBERT\textsubscript{32k} achieves the strongest normalized efficiency score (0.897), substantially outperforming the larger multilingual baselines in terms of performance--efficiency trade-off. The search-cost-aware analysis in Figure~\ref{fig:search_cost_efficiency_six_panel} further illustrates that the DunbaaBERT variants consistently occupy favorable performance--efficiency regions relative to their overall search cost.

Across downstream tasks, the results indicate that increasing the DunbaaBERT vocabulary size from 32k or 52k to 96k does not consistently improve downstream effectiveness under otherwise identical experimental conditions. While DunbaaBERT\textsubscript{96k} often achieves slightly stronger raw classification scores, DunbaaBERT\textsubscript{32k} repeatedly provides the strongest overall efficiency profile. Moreover, although multilingual models such as XLM-R\textsubscript{large} and HPLT\textsubscript{base} can achieve highly competitive predictive performance, the Urdu-specific DunbaaBERT variants generally provide substantially stronger cost-efficiency trade-offs.

\section{Discussion}

The observed trends in UrBLiMP exhibit notable parallels to findings previously reported for Turkish in the SindBERT study~\cite{schmitt-schweter-2026-sindbert}. In particular, SindBERT showed that larger parameter counts and scaling-oriented multilingual approaches do not necessarily translate into consistently superior linguistic competence for morphologically rich languages. Our results point toward similar dynamics for Urdu. Although the 96k DunbaaBERT configuration achieved the lowest pre-training perplexity, this advantage did not consistently transfer to downstream linguistic acceptability performance, where the medium-sized 52k configuration yielded the strongest overall results.

Motivated in part by earlier observations from SindBERT, we placed particular emphasis on corpus quality, diversity, and deduplication during corpus construction; instead of relying solely on raw crawl scale, we deliberately combined multiple heterogeneous Urdu sources and selected CulturaX as a comparatively clean, pre-filtered, and deduplicated web corpus component. Furthermore, the NLLB-derived portion of the corpus underwent additional filtering and processing, reducing the overall corpus size from approximately 22.3GB before deduplication to roughly 17GB in the final training corpus. Taken together, these observations suggest that careful corpus curation, tokenizer construction, and training quality may contribute more strongly to downstream linguistic competence than vocabulary scaling alone.

This interpretation is further supported by the strong performance of HPLT-BERT\textsubscript{ur} and the medium-sized DunbaaBERT\textsubscript{52k} configuration, both of which balance vocabulary coverage with comparatively compact subword inventories. At the same time, the strong HPLT-BERT\textsubscript{ur} results raise additional questions regarding the interaction between corpus composition, pre-training objectives, and linguistic acceptability evaluation. In particular, we speculate that the HPLT corpus construction pipeline may ultimately provide training data that are somewhat better aligned with BLiMP-style grammatical phenomena than our own corpus mixture.

Another open question concerns the role of sentence-level pre-training objectives such as next sentence prediction (NSP). While RoBERTa-style training commonly omits NSP, several strong-performing BERT-family models in morphologically rich languages suggest that discourse-level objectives may still warrant further investigation, particularly for free-word-order and agreement-sensitive phenomena.

Beyond the intrinsic evaluation, the downstream results highlight the effectiveness of the DunbaaBERT models across practical Urdu NLP tasks. Across multiple Urdu classification benchmarks, the DunbaaBERT variants achieved highly competitive results against substantially larger multilingual baselines while consistently maintaining stronger normalized efficiency. In particular, DunbaaBERT\textsubscript{32k} repeatedly provided the strongest overall performance--efficiency trade-off within our experimental setting, indicating that vocabulary scaling alone did not guarantee improved downstream effectiveness in the evaluated models and tasks. Notably, on the USADC benchmark, the DunbaaBERT variants also outperform previously reported GPT-4-based Urdu offensive-language detection results discussed by \citet{maab-etal-2026-prompt}, despite relying on comparatively compact encoder-only architectures. Taken together, our findings indicate that carefully trained Urdu-specific encoder models can remain highly competitive for practical downstream NLP applications, particularly under realistic efficiency and deployment constraints.

More broadly, our findings highlight several promising directions for future Urdu language modeling. While the present work focuses on standard encoder-only architectures with a maximum sequence length of 512 tokens, recent work on long-context transformers suggests that extended-context training may provide additional benefits for discourse-sensitive and document-level phenomena. Architectures such as Longformer~\cite{beltagy2020longformerlongdocumenttransformer}, Nyströmformer~\cite{xiong2021nystromformernystrombasedalgorithmapproximating}, and recent efficient long-context encoder approaches, including ModernBERT~\cite{warner2024smarterbetterfasterlonger}, may therefore represent promising directions for Urdu representation learning, particularly given the prevalence of long-form news articles and document-centric web corpora.

Finally, the current Urdu evaluation landscape still lacks broad standardized downstream benchmark suites. While UrBLiMP provides valuable insight into grammatical acceptability and morphosyntactic competence, broader evaluation settings covering semantic understanding, reasoning, classification, retrieval, and sentence-pair tasks remain comparatively limited. Although GLUE-style Urdu benchmark collections have recently been proposed, we did not include them because the identified resource was available only as an anonymous preprint at the time of this study.\footnote{\url{https://openreview.net/forum?id=ESTMRKI7WK}}

\section{Conclusion}

We introduced DunbaaBERT, a family of Urdu RoBERTa-base models trained with different Byte-BPE vocabulary sizes and evaluated across intrinsic and downstream benchmarks. Overall, the results show that DunbaaBERT is competitive with strong Urdu and multilingual baselines, while the 32k variant often provides the strongest performance--efficiency trade-off. We release the DunbaaBERT models under the MIT license.

\section*{Limitations}

A limitation of the present work is that the downstream evaluation focuses primarily on classification benchmarks. Based on our extensive survey of currently available Urdu resources, these tasks provided the most reliable and reproducible evaluation setting. In contrast, existing Urdu NER datasets often exhibit substantial limitations, including incomplete annotations, limited documentation, or unusually saturated performance levels reported in prior work \cite{10.1145/3329710, zafar-etal-2025-courtroom}, which complicates meaningful comparison and analysis. Furthermore, although GLUE-style Urdu benchmark collections have recently emerged \cite{anonymous2026urduglue}, we did not include them in the present work since the benchmark currently exists only as an anonymous pre-print. Nevertheless, NER and broader benchmark standardization remain important directions for future work, particularly since vocabulary size and tokenization granularity may have stronger effects on sequence labeling tasks than on document-level classification.

In addition, the pre-training corpus was derived primarily from web-based sources and therefore may not fully capture regional, dialectal, or domain-specific variation in Urdu. Like other web-derived corpora, it may also contain social and linguistic biases present in online text.

Finally, although our controlled comparison varies vocabulary size while holding the pre-training corpus, architecture, training budget, and optimization settings constant, we do not report tokenizer-level diagnostics such as subwords per word, vocabulary utilization, or morphology-sensitive segmentation quality. Consequently, our results characterize empirical performance differences among the three vocabulary sizes but do not establish the tokenization mechanisms underlying these differences. Future work should combine such diagnostics with qualitative error analyses to clarify vocabulary- and tokenization-related performance patterns.

\section*{Ethical Considerations}
Like other large-scale language models trained on web-derived corpora, DunbaaBERT may inherit social and linguistic biases present in online text, which can affect downstream applications. In addition, although the training corpus was filtered and deduplicated, large-scale web data may still contain noisy or potentially sensitive content. Finally, pre-training transformer models requires substantial computational resources and therefore contributes to energy consumption and environmental impact.




\section*{Acknowledgments}

The authors dedicate this work in remembrance of the faith of Abraham and in gratitude for the guidance, grace, and mercy that accompanied them throughout this journey.
Further, the authors gratefully acknowledge the scientific support and resources of the AI service infrastructure LRZ AI Systems provided by the Leibniz Supercomputing Centre (LRZ) of the Bavarian Academy of Sciences and Humanities (BAdW), funded by Bayerisches Staatsministerium für Wissenschaft und Kunst (StMWK), and Research and Development Center for Large Language Models (LLMC), National Institute of Informatics, Tokyo.


\bibliography{custom}

\newpage
\appendix

\section{Technical Specifications}
\label{sec:appendix_tech}

\subsection{Computational Setup.}
All experiments were conducted on one GPU compute node. GPU compute node is equipped with NVIDIA A100 GPU with 40GB memory. The models were fine-tuned using PyTorch and HuggingFace Transformers, with mixed-precision training enabled where applicable. 


\subsection{Implementation Specifics}
For all downstream classification benchmarks, we fine-tuned each model under a unified hyperparameter-search setting. We searched over learning rates $\{5\mathrm{e}{-6}, 7\mathrm{e}{-6}, 1\mathrm{e}{-5}, 2\mathrm{e}{-5}, 5\mathrm{e}{-5}\}$ and training batch sizes $\{16, 32, 48, 64\}$ using three random seeds $\{5, 42, 777\}$. Models were trained for up to 30 epochs with early stopping patience of 3, and the best configuration was selected according to validation macro-F1. For final reporting, we evaluated the selected configuration on the held-out test set and report mean $\pm$ standard deviation across seeds. To ensure comparable efficiency measurements, evaluation and inference were performed with a fixed batch size of 8. In addition to accuracy and macro-F1, we also record computational efficiency metrics, including wall-clock training runtime, and inference throughput in samples per second.

For \textsc{UrBLiMP}, we compute accuracy for each category and report the macro-average across all categories as the overall \textsc{UrBLiMP} score. This evaluation captures the models’ ability to distinguish fine-grained grammatical patterns beyond surface-level classification performance. Furthermore, for all other benchmarks, we use an 80/10/10 train/validation/test split and select hyperparameters using validation macro-F1. For COUNT19, a seven-way Urdu news-domain classification dataset, the split is stratified over news categories. For USADC, we use stratified splits over the binary labels, \textit{Normal} and \textit{Abusive}. For the PSL--Kabaddi sentiment benchmark, we stratify the split by sentiment polarity to preserve the distribution of positive, negative, and neutral examples. For the Urdu IMDB dataset, which contains 50,000 balanced positive and negative movie reviews, we likewise use an 80/10/10 split, with the validation set reserved for hyperparameter selection and the held-out test set used only for final reporting. Across all datasets, the reported test results are obtained from the best learning-rate and batch-size configuration selected according to mean validation macro-F1 across the three random seeds. The selected hyperparameters for each model and downstream benchmark are reported in Table~\ref{tab:selected_hyperparameters_downstream}.

\begin{table*}[htb]
\centering
\small
\begin{tabular}{lcccccccc}
\toprule
\multirow{2}{*}{\bfseries Model}
& \multicolumn{2}{c}{\bfseries COUNT19}
& \multicolumn{2}{c}{\bfseries USADC}
& \multicolumn{2}{c}{\bfseries PSL}
& \multicolumn{2}{c}{\bfseries IMDB} \\
\cmidrule(lr){2-3}
\cmidrule(lr){4-5}
\cmidrule(lr){6-7}
\cmidrule(lr){8-9}
& BS & LR & BS & LR & BS & LR & BS & LR \\
\midrule
DunbaaBERT\textsubscript{96k}        & 48 & 1e-5 & 32 & 1e-5 & 16 & 2e-5 & 48 & 2e-5 \\
DunbaaBERT\textsubscript{52k}        & 48 & 1e-5 & 48 & 1e-5 & 64 & 5e-5 & 16 & 2e-5 \\
DunbaaBERT\textsubscript{32k}        & 32 & 5e-6 & 16 & 2e-5 & 32 & 5e-5 & 64 & 5e-5 \\
HPLT-BERT\textsubscript{ur}          & 64 & 5e-6 & 16 & 5e-5 & 64 & 5e-5 & 16 & 1e-5 \\
XLM-R\textsubscript{base}            & 32 & 7e-6 & 16 & 5e-5 & 48 & 5e-5 & 48 & 2e-5 \\
mmBERT\textsubscript{base}           & 32 & 2e-5 & 32 & 5e-5 & 64 & 5e-5 & 32 & 2e-5 \\
mBERT\textsubscript{base}            & 16 & 7e-6 & 32 & 5e-5 & 16 & 2e-5 & 48 & 7e-6 \\
mmBERT\textsubscript{small}          & 64 & 5e-5 & 32 & 2e-5 & 48 & 2e-5 & 64 & 5e-5 \\
Urdu-RoBERTa\textsubscript{small}    & 48 & 5e-5 & 32 & 5e-5 & 48 & 2e-5 & 32 & 5e-6 \\
XLM-R\textsubscript{large}           & 16 & 5e-6 & 48 & 5e-5 & 48 & 2e-5 & 64 & 5e-6 \\
\bottomrule
\end{tabular}
\caption{Hyperparameters of the best downstream task models for each task and pre-trained model. BS denotes batch size and LR denotes learning rate. For each dataset, the reported configuration is selected using the highest mean validation macro-F1 across seed runs.}
\label{tab:selected_hyperparameters_downstream}
\end{table*}

\section{Qualitative Analysis of Heuristic Penalties}
\label{sec:heuristic_penalties}

To better understand the behavior of the proposed heuristic filtering approach, Table~\ref{tab:worst5_top5} presents the five highest-ranked and five lowest-ranked candidates according to the final scoring function. These examples provide a qualitative sanity check illustrating both which types of textual fragments are preserved and which receive the strongest penalties during corpus construction.

The lowest-ranked candidates are dominated by typical web-crawling artifacts, including truncated navigation snippets, ``Read more'' fragments, timestamps, category labels, newsroom metadata, and malformed or noisy Unicode remnants. Interestingly, several of these examples still achieve moderately high semantic similarity scores despite being linguistically uninformative. This observation suggests that embedding-based similarity alone is insufficient for reliable corpus filtering in large-scale crawl-based settings.

In contrast, the highest-ranked candidates largely consist of coherent and informative Urdu text fragments, including encyclopedic descriptions and geographically grounded narrative content. Although minor traces of noise or mixed-language artifacts remain present in some examples, the overall linguistic quality is substantially higher than in the heavily penalized candidates.

Instead, the results indicate that the additional heuristic penalty component successfully suppresses non-linguistic boilerplate and template artifacts that would otherwise remain in the final corpus. The examples further illustrate the importance of combining semantic scoring with lightweight rule-based filtering during corpus curation for morphologically rich low-resource languages such as Urdu.

\begin{table*}[t]
\centering
\small

\begin{tabular}{c}
\toprule
\makebox[0.98\textwidth][l]{%
\hspace{0.4cm}\textbf{Final}%
\hspace{0.7cm}\textbf{Sim.}%
\hspace{0.7cm}\textbf{Pen.}%
\hspace{0.7cm}\textbf{Text}%
} \\
\midrule
\includegraphics[width=0.98\textwidth]{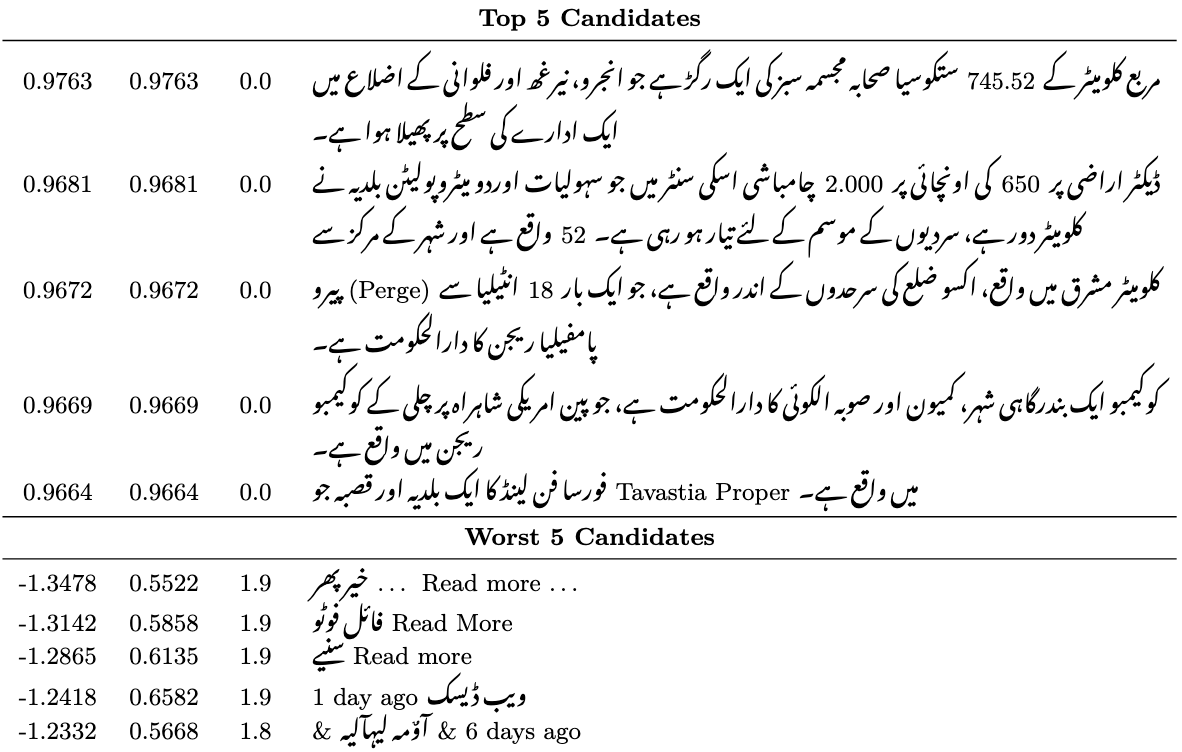} \\
\bottomrule
\end{tabular}

\vspace{0.3em}

\caption{Qualitative comparison between the five highest-ranked candidates (top) and five lowest-ranked candidates (bottom) according to the proposed heuristic scoring function. The columns report the final heuristic score (Final), semantic similarity score (Sim.), penalty value (Pen.), and candidate sentence (Text).}
\label{tab:worst5_top5}

\end{table*}

\section{Model Properties}
\label{sec:model_properties}
Table~\ref{tab:params_he} summarizes the vocabulary sizes and parameter counts of the Urdu and multilingual models considered in our evaluation.
Urdu-RoBERTa\textsubscript{small} (126M) and HPLT\textsubscript{base} (124M) represent strong monolingual Urdu baselines, while mmBERT\textsubscript{small} (140M) and mmBERT\textsubscript{base} (307M) provide multilingual comparison points specifically designed for massively multilingual settings.

Our DunbaaBERT variants systematically investigate the impact of vocabulary size on model scaling and downstream performance.
DunbaaBERT\textsubscript{32k} contains 111M parameters and uses a compact 32k-token vocabulary, while DunbaaBERT\textsubscript{52k}\textsubscript{base} (126M) closely matches the parameter scale of Roberta-Urdu and HPLT.
The larger DunbaaBERT\textsubscript{96k}\textsubscript{base} increases the vocabulary size to 96k tokens, resulting in a substantially larger parameter count of 160M.

For multilingual points of reference, mBERT contains 178M parameters with a WordPiece vocabulary of approximately 120k tokens, while XLM-R\textsubscript{base} and XLM-R\textsubscript{large} contain 278M and 560M parameters, respectively, using 250k-token SentencePiece vocabularies.
All values were extracted using Huggingface’s \texttt{transformers} library.

\begin{table}[H]

\centering\small
\begin{tabular}{lcc}%
    \hline
    \bfseries Model & \bfseries Vocab Size & \bfseries \#Params \\
    \hline
    \csvreader[late after line = \\]{params.csv}{}%
     {\csvcoli & \csvcolii & \csvcoliii}
    \hline
\end{tabular}
\caption{Vocabulary size and total parameter count for Urdu transformer-based models.  
Values were extracted using Huggingface’s \texttt{transformers} library.}\label{tab:params_he}
\end{table}

\section{Efficiency Performance}
\label{sec:efficiency_performance}

\subsection{Predictive Performance and Inference Throughput}

\begin{figure*}[t]
    \centering
    \includegraphics[width=0.82\textwidth]{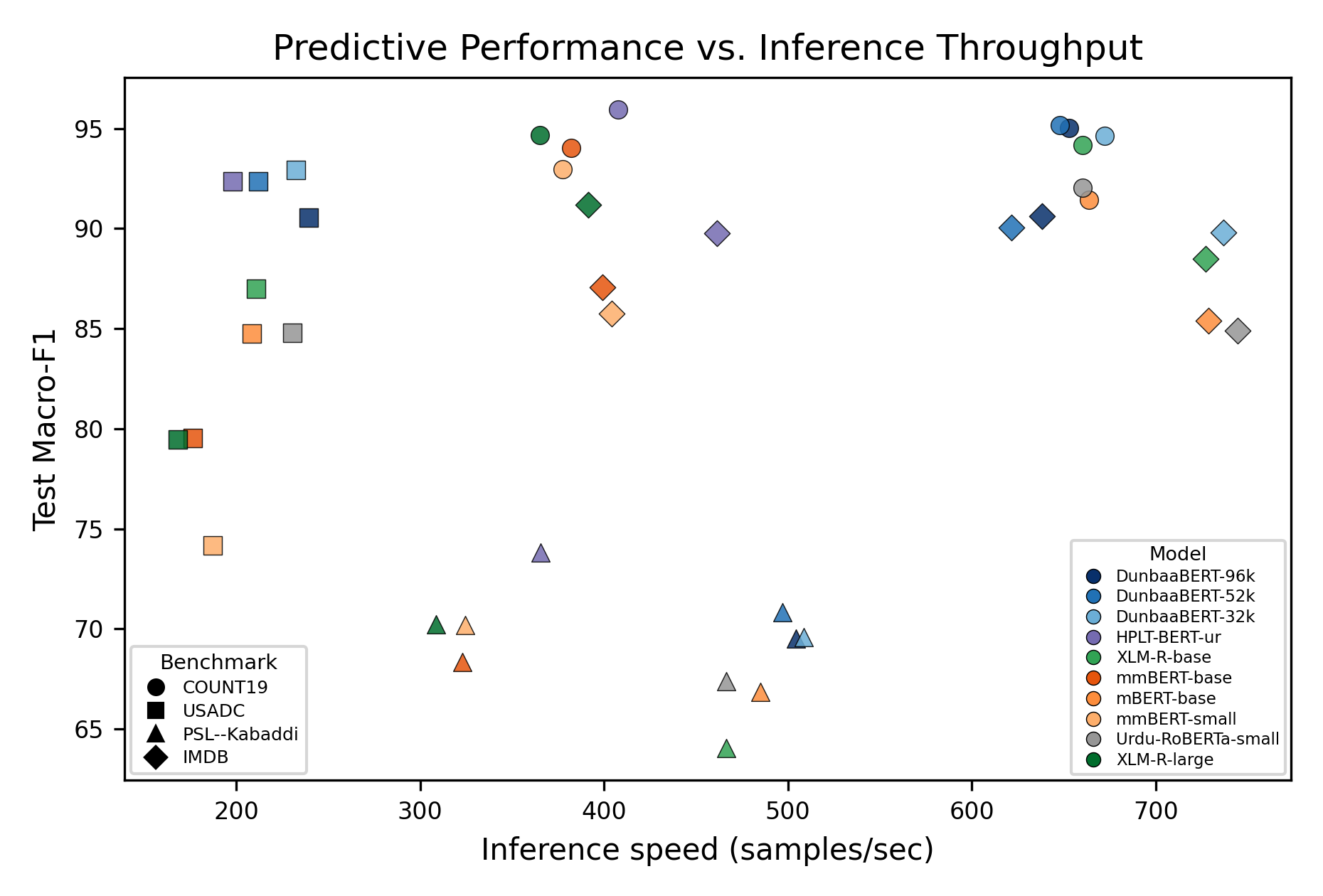}
    \caption{
    Predictive performance (test Macro-F1) versus inference throughput (samples per second) across downstream Urdu benchmarks. For each model and benchmark, results are based on the selected HPO configuration over learning rates and batch sizes, evaluated across three random seeds with early stopping patience 3. Marker shapes denote benchmark datasets, and colors denote model families.
    }
    \label{fig:macro_f1_inference_speed}
\end{figure*}

Figure~\ref{fig:macro_f1_inference_speed} shows the relationship between predictive performance and inference throughput across the evaluated Urdu benchmarks. Unlike reporting macro-F1 alone, this analysis highlights whether models maintain strong performance while also supporting faster inference. The upper-right region is the most desirable, indicating both high test macro-F1 and high samples-per-second throughput. Across tasks, the DunbaaBERT variants frequently occupy this favorable region, suggesting that the proposed Urdu-specific models provide not only competitive performance across all tasks but also efficient inference behavior compared with multilingual and other Urdu baselines.

\subsection{Downstream Training and Inference Efficiency}
\label{sec:downstream_treaining_inference_efficiency}

\begin{figure*}[t]

    \centering

    \includegraphics[width=0.82\textwidth]{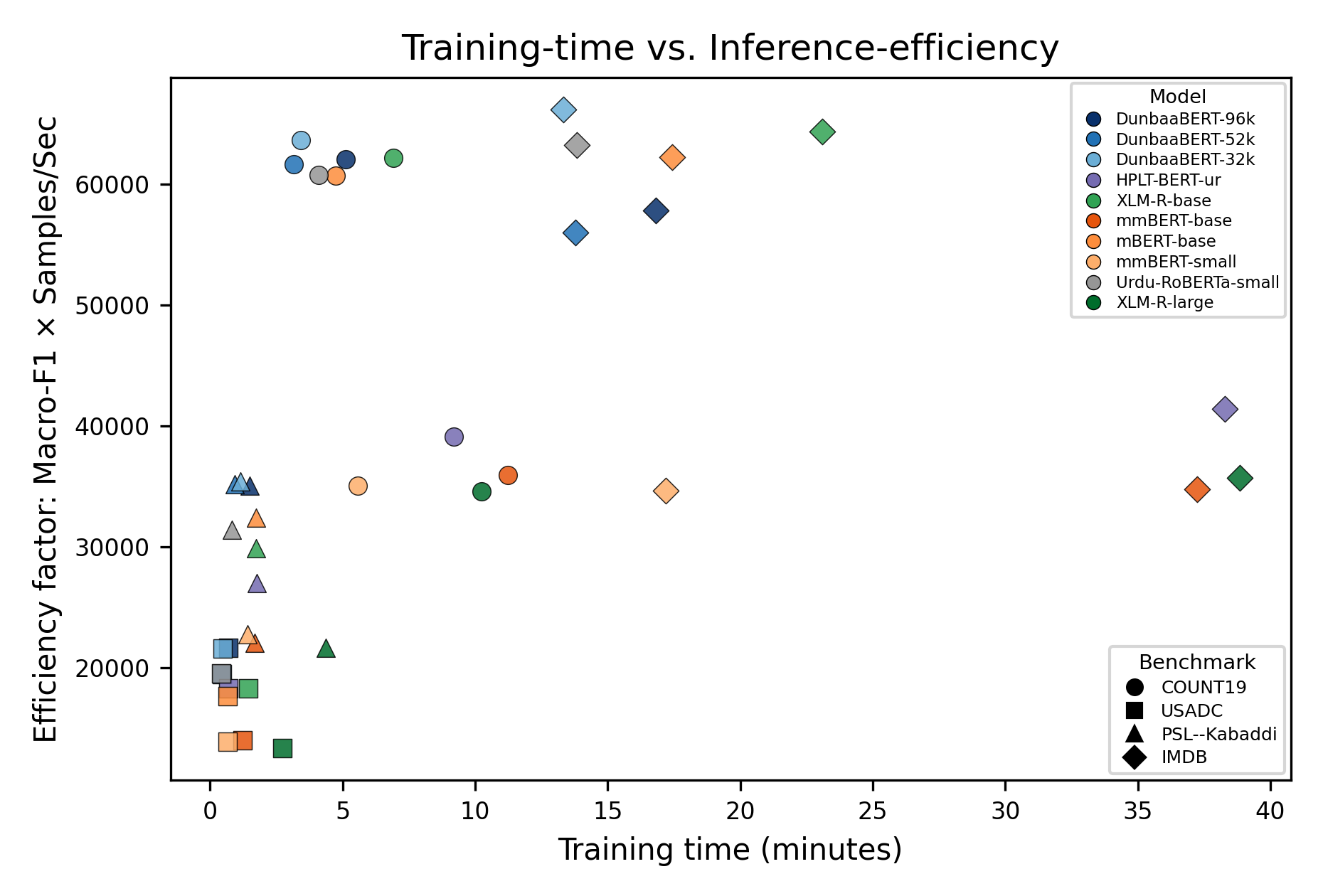}

    \caption{Training-time versus inference-efficiency trade-off across downstream Urdu benchmarks. The x-axis shows average fine-tuning time, and the y-axis shows the efficiency factor, defined as test Macro-F1 $\times$ inference samples per second. Results are based on the best HPO configuration across learning rates and batch sizes, averaged over three seeds with early stopping patience 3. Marker shapes indicate benchmarks, and colors indicate model families.
    }

    \label{fig:training_time_efficiency_tradeoff}

\end{figure*}

Figure~\ref{fig:training_time_efficiency_tradeoff} analyzes the efficiency--performance trade-off by combining downstream predictive quality and inference throughput into a single efficiency factor, defined as macro-F1 $\times$ inference samples per second. The plot complements the numerical results by showing how models compare not only in terms of final task performance, but also with respect to the downstream fine-tuning cost. All points are based on the best hyperparameter configurations selected through validation performance under the same HPO setting, with early stopping using patience 3. A model is preferable when it achieves a higher efficiency factor with lower training time, corresponding to the upper-left region of the plot. The DunbaaBERT variants consistently appear in this favorable region across multiple benchmarks, indicating that they converge efficiently during downstream training while maintaining strong inference efficiency.

\subsection{Performance--Efficiency Across all 60 Configurations}
\label{sec:performance_efficiency_all_60}

To assess whether the models are consistently efficient beyond a single best run, Table~\ref{tab:all_benchmark_normalized_efficiency} reports average results over 60 hyperparameter configurations for each model and benchmark, corresponding to $5$ learning rates $\times$ $4$ batch sizes $\times$ $3$ random seeds. This analysis combines predictive quality, inference speed, and downstream training time under the same early-stopping setting. Overall, the DunbaaBERT variants show strong performance–efficiency trade-offs across benchmarks, often achieving the highest or second-highest normalized efficiency while maintaining competitive Macro-F1 and relatively low fine-tuning time. This suggests that the Urdu-specific DunbaaBERT models are not only effective in terms of task performance, but also efficient across a broad range of training configurations.

\begin{table*}[t]
\centering
\small
\resizebox{\textwidth}{!}{%
\begin{tabular}{llcccc}
\toprule
\textbf{Benchmark} & \textbf{Model} & \textbf{Macro-F1} & \textbf{SPS} & \textbf{Train Time (min)} & \textbf{Norm. Eff.} \\
\midrule

\multirow{10}{*}{COUNT19}
& DunbaaBERT\textsubscript{96k} & \underline{94.90$_{\pm0.50}$} & 645.34$_{\pm14.48}$ & 3.88$_{\pm1.37}$ & 0.916 \\
& DunbaaBERT\textsubscript{52k} & 94.86$_{\pm0.48}$ & 654.38$_{\pm9.73}$ & \textbf{3.18$_{\pm0.85}$} & \underline{0.929} \\
& DunbaaBERT\textsubscript{32k} & 94.44$_{\pm0.58}$ & \textbf{668.26$_{\pm10.59}$} & 3.33$_{\pm0.79}$ & \textbf{0.944} \\
& HPLT-BERT\textsubscript{ur} & \textbf{95.67$_{\pm0.42}$} & 413.68$_{\pm8.78}$ & 8.80$_{\pm2.43}$ & 0.592 \\
& XLM-R\textsubscript{base} & 93.18$_{\pm0.86}$ & 656.25$_{\pm10.61}$ & 5.72$_{\pm1.97}$ & 0.915 \\
& mmBERT\textsubscript{base} & 93.12$_{\pm1.02}$ & 382.11$_{\pm6.38}$ & 7.17$_{\pm2.35}$ & 0.532 \\
& mBERT\textsubscript{base} & 90.57$_{\pm0.86}$ & 635.04$_{\pm24.88}$ & 4.65$_{\pm1.35}$ & 0.861 \\
& mmBERT\textsubscript{small} & 92.54$_{\pm0.76}$ & 382.23$_{\pm5.70}$ & 4.99$_{\pm1.58}$ & 0.529 \\
& Urdu-RoBERTa\textsubscript{small} & 91.74$_{\pm0.54}$ & \underline{663.41$_{\pm8.70}$} & \underline{3.31$_{\pm0.94}$} & 0.911 \\
& XLM-R\textsubscript{large} & 93.92$_{\pm1.12}$ & 365.58$_{\pm4.26}$ & 10.25$_{\pm3.69}$ & 0.514 \\

\midrule
\multirow{10}{*}{USADC}
& DunbaaBERT\textsubscript{96k} & 80.11$_{\pm20.21}$ & \textbf{236.71$_{\pm19.24}$} & 0.54$_{\pm0.24}$ & 0.801 \\
& DunbaaBERT\textsubscript{52k} & \textbf{86.23$_{\pm14.70}$} & \underline{235.66$_{\pm25.42}$} & \underline{0.45$_{\pm0.16}$} & \textbf{0.858} \\
& DunbaaBERT\textsubscript{32k} & 72.37$_{\pm25.03}$ & 230.43$_{\pm10.86}$ & \textbf{0.36$_{\pm0.15}$} & 0.704 \\
& HPLT-BERT\textsubscript{ur} & 84.17$_{\pm12.04}$ & 197.72$_{\pm9.04}$ & 0.65$_{\pm0.24}$ & 0.703 \\
& XLM-R\textsubscript{base} & 68.13$_{\pm19.34}$ & 223.20$_{\pm18.77}$ & 1.10$_{\pm0.52}$ & 0.642 \\
& mmBERT\textsubscript{base} & 80.18$_{\pm4.35}$ & 178.84$_{\pm7.26}$ & 1.27$_{\pm0.30}$ & 0.606 \\
& mBERT\textsubscript{base} & 71.42$_{\pm12.32}$ & 210.24$_{\pm11.51}$ & 0.69$_{\pm0.22}$ & 0.634 \\
& mmBERT\textsubscript{small} & 71.30$_{\pm7.99}$ & 184.14$_{\pm10.21}$ & 0.70$_{\pm0.21}$ & 0.555 \\
& Urdu-RoBERTa\textsubscript{small} & \underline{85.60$_{\pm3.14}$} & 224.36$_{\pm9.35}$ & 0.54$_{\pm0.15}$ & \underline{0.811} \\
& XLM-R\textsubscript{large} & 71.32$_{\pm13.32}$ & 165.80$_{\pm8.45}$ & 2.14$_{\pm0.73}$ & 0.500 \\

\midrule
\multirow{10}{*}{PSL--Kabaddi}
& DunbaaBERT\textsubscript{96k} & 62.54$_{\pm5.44}$ & \underline{498.79$_{\pm16.09}$} & 0.97$_{\pm0.43}$ & \underline{0.593} \\
& DunbaaBERT\textsubscript{52k} & 62.61$_{\pm5.80}$ & 498.10$_{\pm10.85}$ & \underline{0.88$_{\pm0.38}$} & 0.592 \\
& DunbaaBERT\textsubscript{32k} & 63.09$_{\pm5.85}$ & \textbf{526.39$_{\pm11.10}$} & \textbf{0.83$_{\pm0.39}$} & \textbf{0.631} \\
& HPLT-BER\textsubscript{ur} & 63.00$_{\pm7.23}$ & 364.45$_{\pm6.36}$ & 1.33$_{\pm0.50}$ & 0.436 \\
& XLM-R\textsubscript{base} & 57.01$_{\pm3.89}$ & 474.34$_{\pm7.64}$ & 1.87$_{\pm0.66}$ & 0.514 \\
& mmBERT\textsubscript{base} & \textbf{69.30$_{\pm3.90}$} & 320.04$_{\pm7.02}$ & 1.86$_{\pm0.48}$ & 0.421 \\
& mBERT\textsubscript{base} & 58.34$_{\pm7.44}$ & 478.44$_{\pm18.21}$ & 1.48$_{\pm0.60}$ & 0.530 \\
& mmBERT\textsubscript{small} & \underline{68.44$_{\pm3.45}$} & 325.15$_{\pm4.50}$ & 1.40$_{\pm0.40}$ & 0.423 \\
& Urdu-RoBERTa\textsubscript{small} & 62.82$_{\pm7.73}$ & 477.33$_{\pm17.86}$ & 1.03$_{\pm0.43}$ & 0.570 \\
& XLM-R\textsubscript{large} & 63.38$_{\pm9.06}$ & 312.24$_{\pm7.13}$ & 3.60$_{\pm1.42}$ & 0.376 \\

\midrule
\multirow{10}{*}{IMDB Urdu}
& DunbaaBERT\textsubscript{96k} & \underline{90.19$_{\pm0.38}$} & 633.91$_{\pm19.91}$ & 13.51$_{\pm5.74}$ & 0.771 \\
& DunbaaBERT\textsubscript{52k} & 90.05$_{\pm0.31}$ & 672.41$_{\pm63.26}$ & 11.76$_{\pm3.59}$ & 0.817 \\
& DunbaaBERT\textsubscript{32k} & 89.76$_{\pm0.29}$ & 735.47$_{\pm10.79}$ & \textbf{11.18$_{\pm3.69}$} & \textbf{0.890} \\
& HPLT-BERT\textsubscript{ur} & 89.47$_{\pm0.29}$ & 456.80$_{\pm14.47}$ & 39.14$_{\pm12.06}$ & 0.551 \\
& XLM-R\textsubscript{base} & 88.35$_{\pm0.35}$ & 722.30$_{\pm11.15}$ & 15.97$_{\pm5.52}$ & \underline{0.861} \\
& mmBERT\textsubscript{base} & 86.74$_{\pm0.41}$ & 404.76$_{\pm5.07}$ & 20.55$_{\pm6.12}$ & 0.474 \\
& mBERT\textsubscript{base} & 85.05$_{\pm0.70}$ & \underline{737.58$_{\pm8.61}$} & 13.49$_{\pm5.04}$ & 0.846 \\
& mmBERT\textsubscript{small} & 85.08$_{\pm0.58}$ & 404.45$_{\pm4.30}$ & 12.44$_{\pm4.15}$ & 0.464 \\
& Urdu-RoBERTa\textsubscript{small} & 84.47$_{\pm0.55}$ & \textbf{741.35$_{\pm8.07}$} & \underline{11.48$_{\pm3.34}$} & 0.845 \\
& XLM-R\textsubscript{large} & \textbf{90.52$_{\pm1.02}$} & 391.52$_{\pm4.64}$ & 26.18$_{\pm9.45}$ & 0.478 \\

\bottomrule
\end{tabular}%
}
\caption{
For each model and benchmark, results are summarized over all 60 unique configurations, corresponding to $5$ learning rates $\times$ $4$ batch sizes $\times$ $3$ random seeds, with early stopping using patience 3. Test Macro-F1, inference throughput in samples per second (SPS), and downstream fine-tuning time are reported as mean$_{\pm}$standard deviation. Norm. Eff. denotes normalized efficiency, computed within each benchmark from the corresponding mean Macro-F1 and mean SPS as $(\mathrm{Macro\mbox{-}F1}/100) \times (\mathrm{SPS}/\max(\mathrm{SPS}))$. Higher values indicate a stronger performance--efficiency trade-off. Best results are shown in bold and second-best results are underlined.
}
\label{tab:all_benchmark_normalized_efficiency}
\end{table*}

\subsection{Search-Cost--Aware Performance and Efficiency}
\label{sec:search_cost_aware_performance_efficiency}

\begin{figure*}[t]
\centering

\begin{subfigure}[t]{0.48\textwidth}
    \centering
    \includegraphics[width=\linewidth]{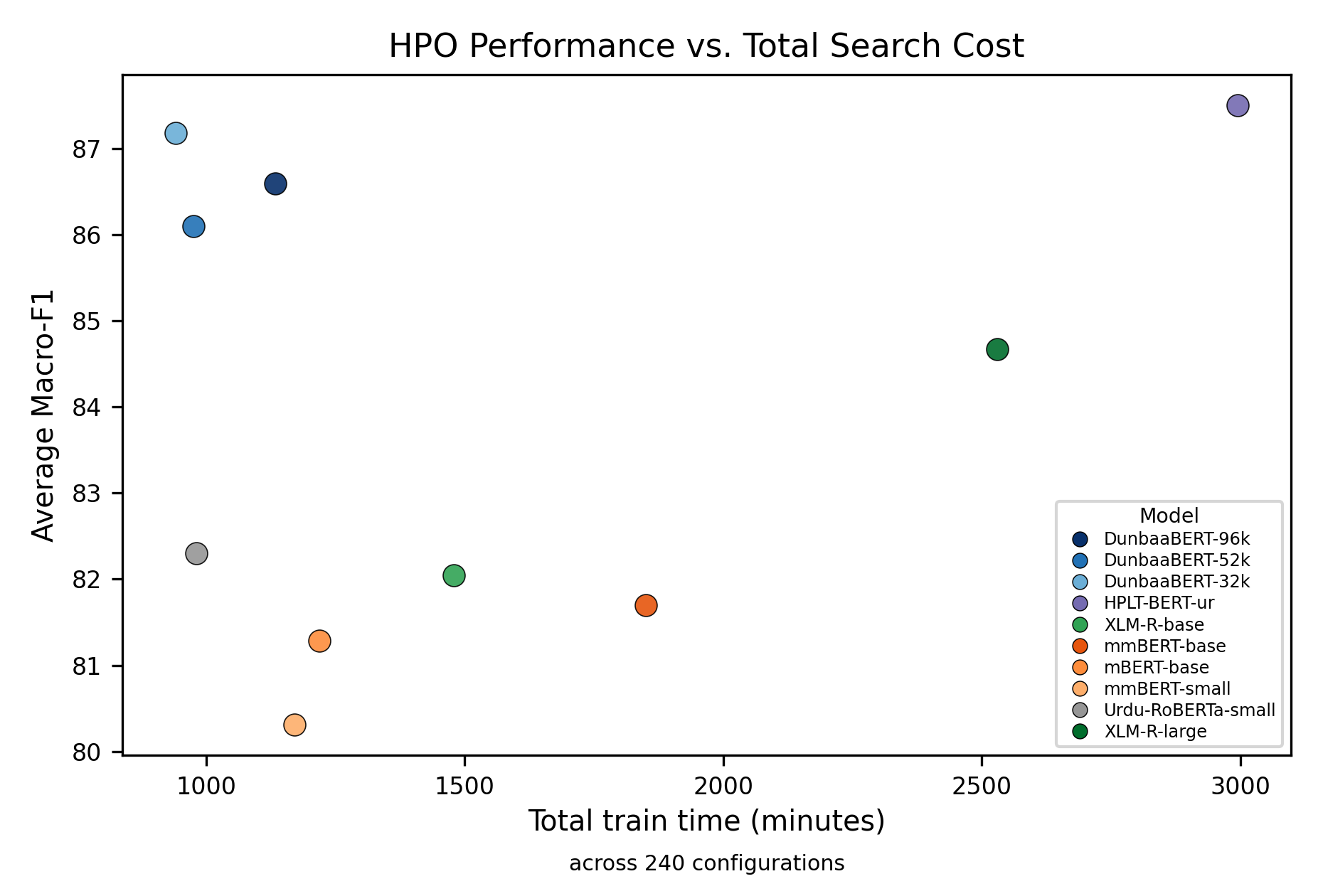}
    \caption{HPO-selected Macro-F1 versus total search cost.}
    \label{fig:hpo_search_cost_macro_f1}
\end{subfigure}
\hfill
\begin{subfigure}[t]{0.48\textwidth}
    \centering
    \includegraphics[width=\linewidth]{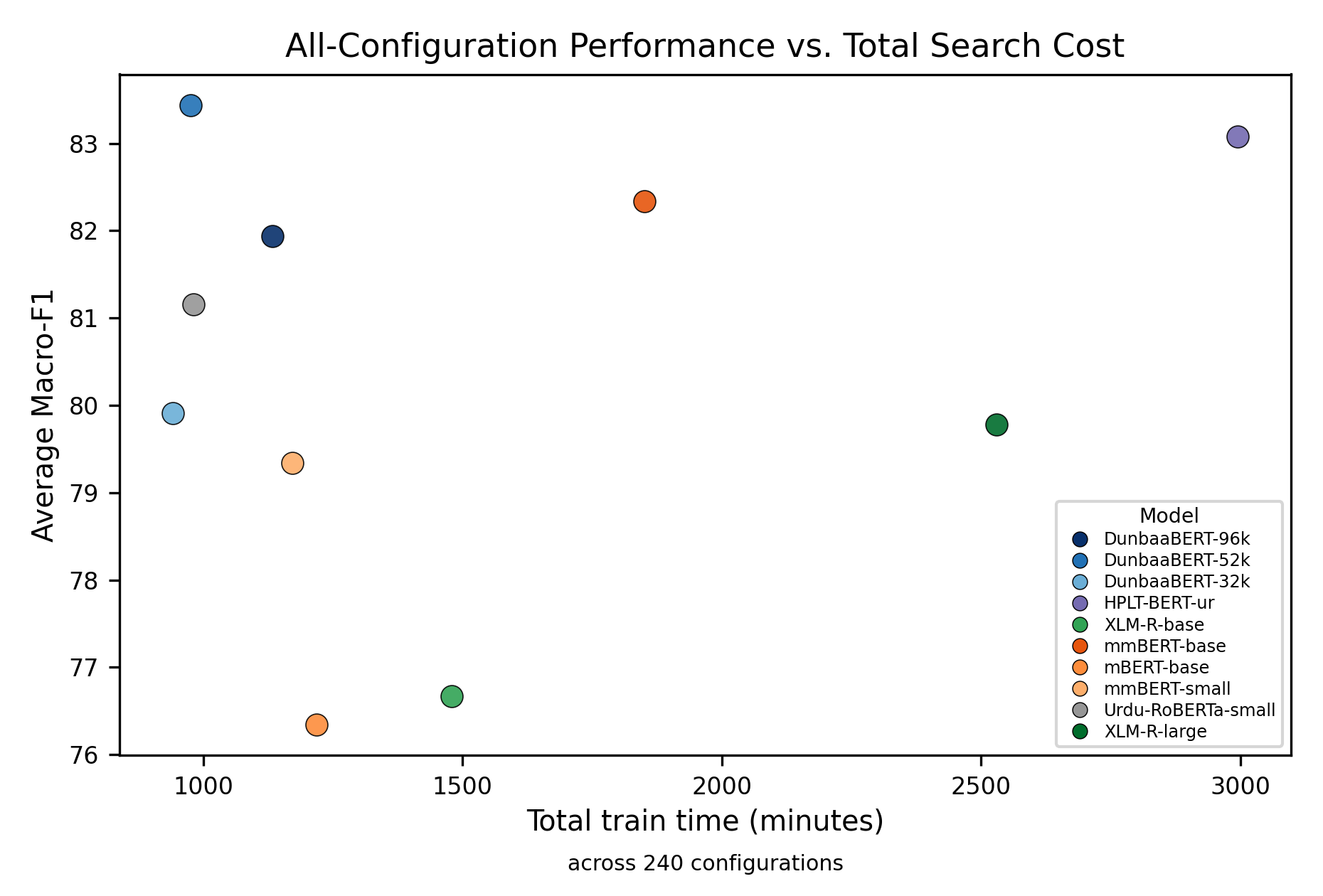}
    \caption{All-configuration Macro-F1 versus total search cost.}
    \label{fig:allconfig_search_cost_macro_f1}
\end{subfigure}

\vspace{0.6em}

\begin{subfigure}[t]{0.48\textwidth}
    \centering
    \includegraphics[width=\linewidth]{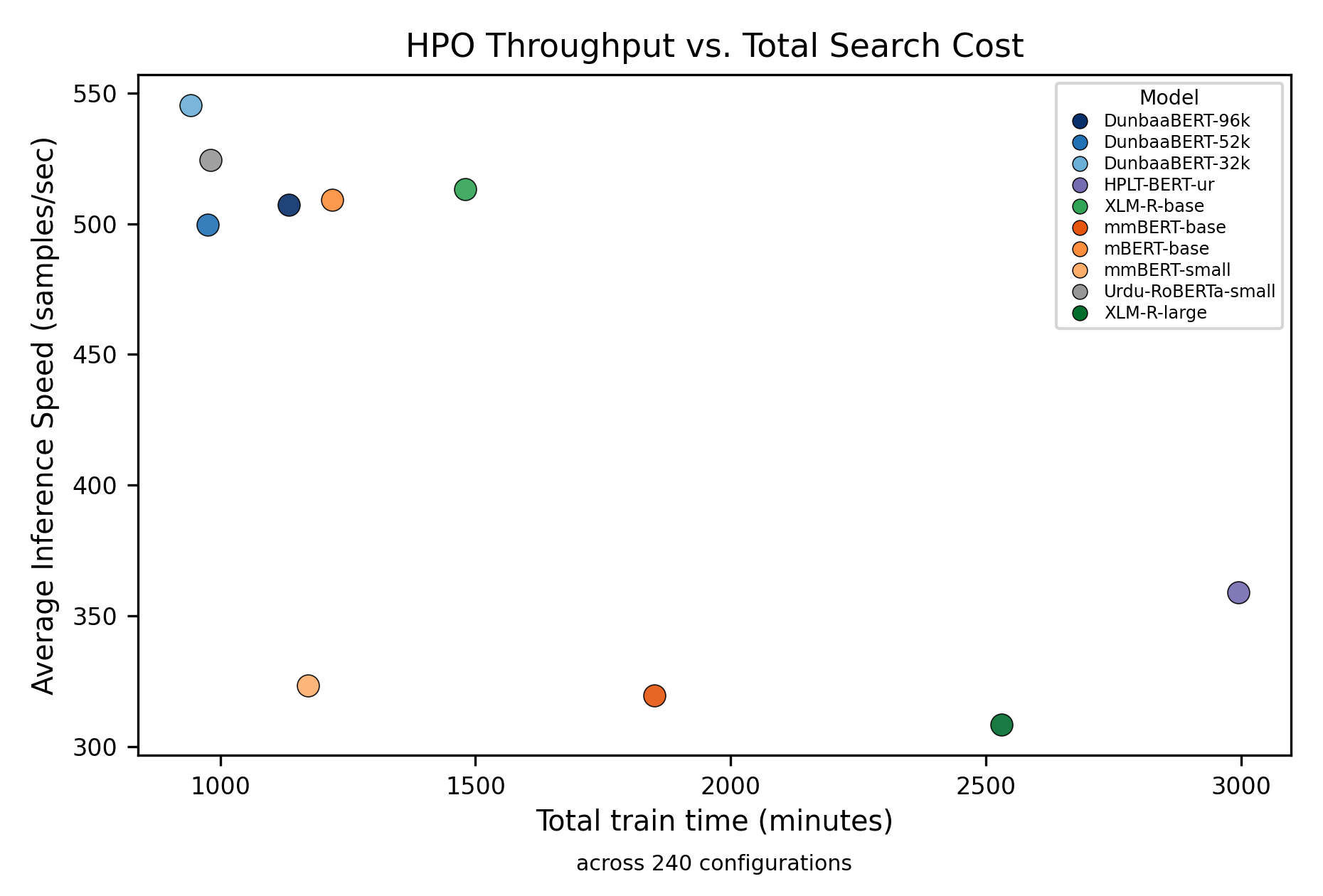}
    \caption{HPO-selected inference throughput versus total search cost.}
    \label{fig:hpo_search_cost_sps}
\end{subfigure}
\hfill
\begin{subfigure}[t]{0.48\textwidth}
    \centering
    \includegraphics[width=\linewidth]{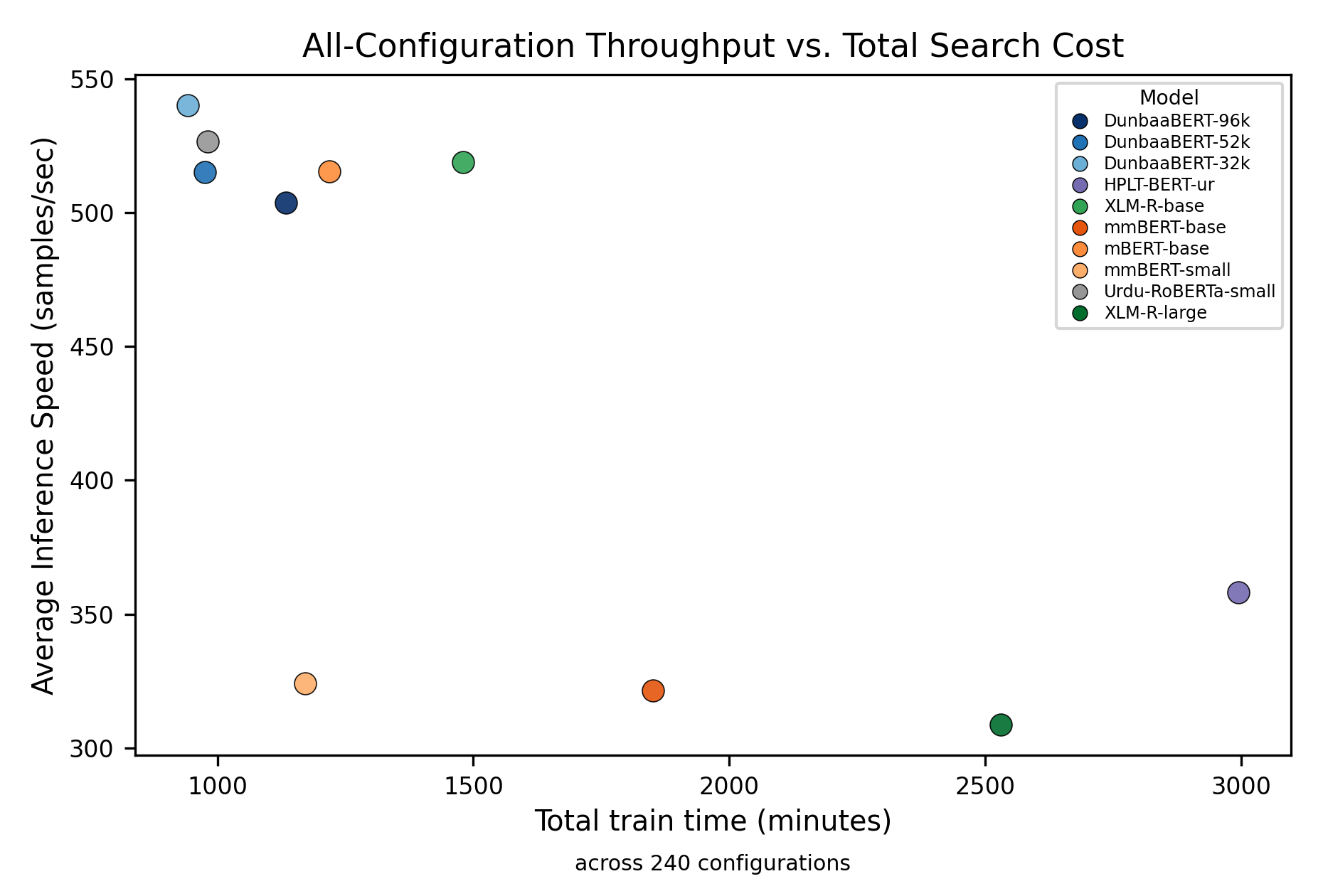}
    \caption{All-configuration inference throughput versus total search cost.}
    \label{fig:allconfig_search_cost_sps}
\end{subfigure}

\vspace{0.6em}

\begin{subfigure}[t]{0.48\textwidth}
    \centering
    \includegraphics[width=\linewidth]{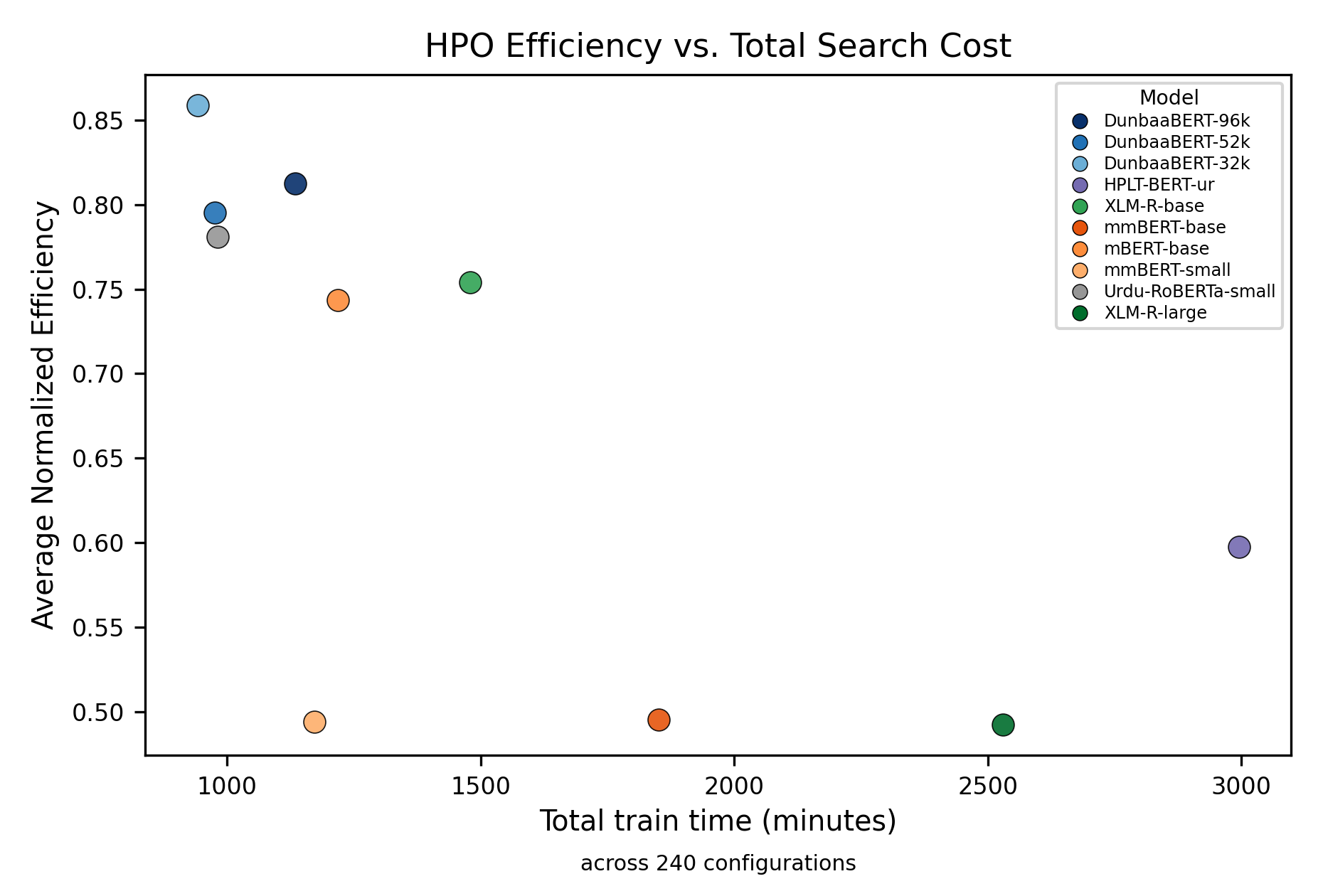}
    \caption{HPO-selected normalized efficiency versus total search cost.}
    \label{fig:hpo_search_cost_norm_eff}
\end{subfigure}
\hfill
\begin{subfigure}[t]{0.48\textwidth}
    \centering
    \includegraphics[width=\linewidth]{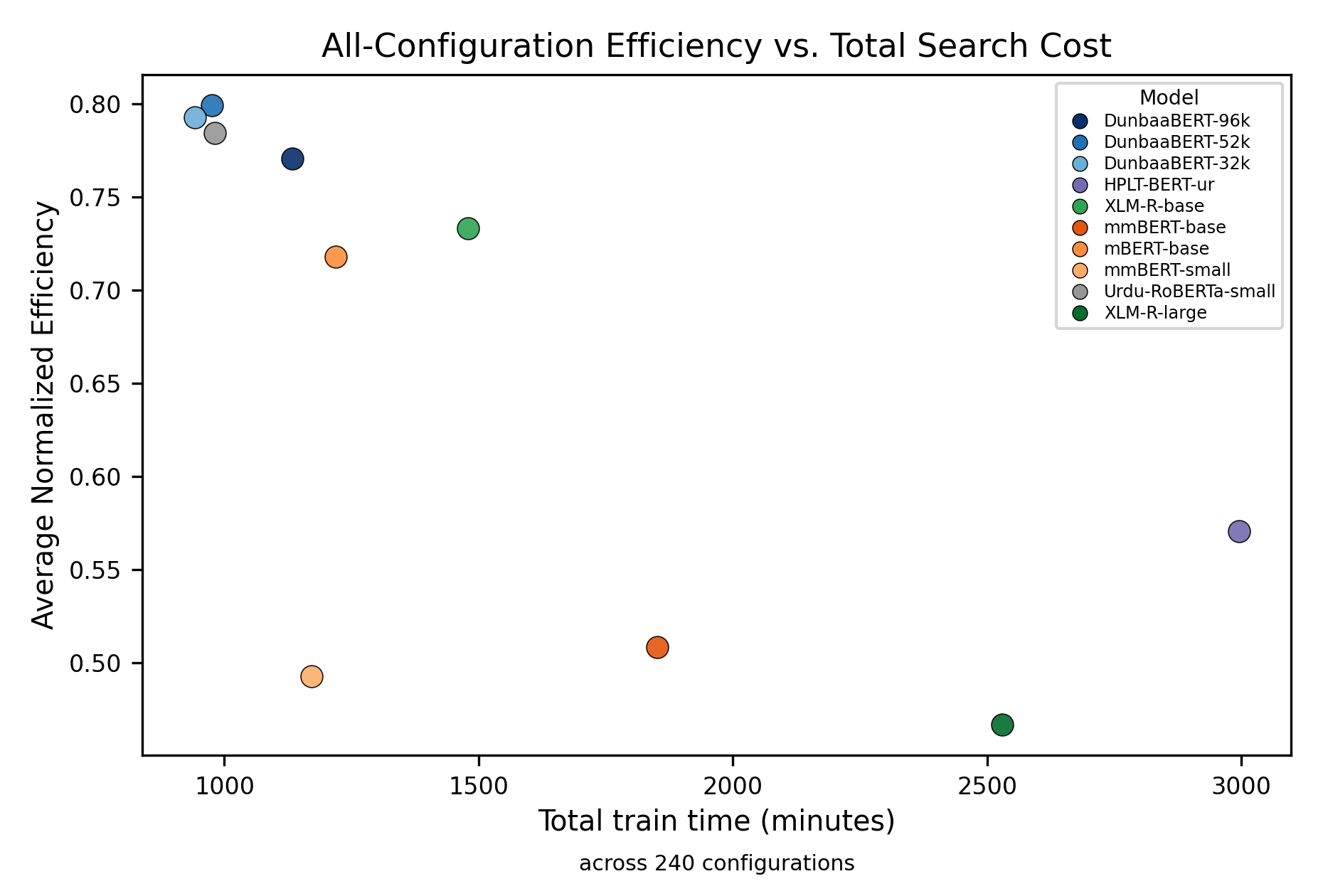}
    \caption{All-configuration normalized efficiency versus total search cost.}
    \label{fig:allconfig_search_cost_norm_eff}
\end{subfigure}

\caption{
Aggregate relationship between total hyperparameter-search training cost and model performance--efficiency behavior across four downstream Urdu benchmarks (COUNT19, USADC, PSL--Kabbadi, IMDB-Urdu). Each point represents one model, and the x-axis reports total training time over the full unique search grid of $5$ learning rates $\times$ $4$ batch sizes $\times$ $3$ seeds across $4$ benchmarks, i.e., $240$ configurations per model. The left column reports results after selecting the best HPO setting, while the right column summarizes behavior over all configurations. Panels (a,b) compare predictive performance using test Macro-F1, panels (c,d) compare inference throughput in samples per second, and panels (e,f) compare normalized efficiency (Norm. Eff.) \ref{subsec:efficiency_metric}. Points closer to the upper-left region indicate a stronger trade-off between effectiveness and search cost.
}
\label{fig:search_cost_efficiency_six_panel}
\end{figure*}

Figure~\ref{fig:search_cost_efficiency_six_panel} further analyzes the relationship between total hyperparameter search cost and model effectiveness. Each point represents one model, where the x-axis reports the total training time required to evaluate the full search space across the four benchmarks, corresponding to $60$ configurations per benchmark and $240$ configurations per model. The left column reports scores from the HPO-selected setting, while the right column reports the corresponding all-configuration summaries.

Overall, the DunbaaBERT variants provide a strong search-cost--performance trade-off. In the HPO-selected setting, DunbaaBERT\textsubscript{32k} achieves the most favorable balance, combining high average Macro-F1, strong inference throughput, and the highest normalized efficiency while requiring relatively low total search cost. DunbaaBERT\textsubscript{96k} also performs competitively, particularly in Macro-F1 and normalized efficiency, but with slightly higher search cost. In contrast, HPLT-BERT\textsubscript{ur} obtains strong HPO-selected Macro-F1 but requires substantially larger search time, weakening its overall cost-effectiveness.

The all-configuration results show a slightly different trend. DunbaaBERT\textsubscript{52k} appears particularly robust when averaging over the full configuration space, achieving strong Macro-F1 and normalized efficiency at relatively low search cost. This suggests that while DunbaaBERT\textsubscript{32k} is highly effective after HPO selection, DunbaaBERT\textsubscript{52k} may be more stable across broader hyperparameter settings. Larger multilingual models such as XLM-R\textsubscript{large} and HPLT-BERT\textsubscript{ur} generally require higher search cost, and their efficiency gains are less consistent. These results indicate that Urdu-specific pretraining can improve not only downstream accuracy but also the practical cost-efficiency of model selection.

\section{Computational Cost of Benchmarking}
\label{sec:computational_cost_of_benchmarking}

In addition to predictive performance and inference efficiency, we report the total downstream fine-tuning cost required for the benchmark suite. Table~\ref{tab:computation_time_by_benchmark} summarizes the cumulative wall-clock training time across all evaluated configurations. For each benchmark, this includes 600 unique fine-tuning runs, corresponding to 10 models, 5 learning rates, 4 batch sizes, and 3 random seeds. Overall, the full evaluation required 2400 fine-tuning runs and 254 hours and 38 minutes of cumulative training time. This highlights the computational cost of systematic benchmarking and motivates reporting efficiency-oriented metrics alongside task performance.

\begin{table}[H]
\centering
\small
\begin{tabular}{lccc}
\toprule
\textbf{Benchmark} & \textbf{Configs} & \textbf{Computation Time} & \textbf{Days} \\
\midrule
COUNT19 & 600 & 55:15 & 2.30 \\
USADC & 600 & 8:26 & 0.35 \\
PSL--Kabaddi & 600 & 15:15 & 0.64 \\
IMDB Urdu & 600 & 175:42 & 7.32 \\
\midrule
\textbf{Total} & \textbf{2400} & \textbf{254:38} & \textbf{10.61} \\
\bottomrule
\end{tabular}
\caption{
Total downstream fine-tuning computation time across benchmarks. For each benchmark, computation time is summed over 600 unique fine-tuning runs, corresponding to 10 models $\times$ 5 learning rates $\times$ 4 batch sizes $\times$ 3 random seeds. Thus, each individual model contributes 60 configurations per benchmark. Computation Time is reported in hours:minutes and Days gives the equivalent duration in days. 
}
\label{tab:computation_time_by_benchmark}
\end{table}

\end{document}